\documentclass{aamas2009}

\usepackage{amscd,amsmath,graphicx,latexsym}
\usepackage{cite,setspace,url,xspace}
\newtheorem{theorem}{Theorem}
\newtheorem{example}{Example}

\TitleForCitationInfo{Manipulation and gender neutrality in stable marriage procedures}
\AuthorsForCitationInfo{M.S.~Pini, F.~Rossi, K.B.~Venable, T.~Walsh}

\title{Manipulation and gender neutrality in stable marriage procedures
}

\numberofauthors{4} 
\author{
\alignauthor
Maria Silvia Pini\\
       \affaddr{Univ. of Padova, Padova, Italy}\\
       \email{mpini@math.unipd.it}
\alignauthor
Francesca Rossi\\
       \affaddr{Univ. of Padova, Padova, Italy}\\
       \email{frossi@math.unipd.it}
\alignauthor
K. Brent Venable\\
       \affaddr{Univ. of Padova, Padova, Italy}\\
       \email{kvenable@math.unipd.it}
\and
\alignauthor
Toby Walsh\\
       \affaddr{NICTA and UNSW, Sydney, Australia}\\
       \email{toby.walsh@nicta.com.au}
}

\begin{document}


\maketitle
\begin{abstract}

The stable marriage problem is a well-known problem
of matching men to women so that no man and woman who
are not married to each other both prefer each other. 
Such a problem has a wide variety of
practical applications ranging from matching resident doctors to
hospitals to matching students to schools.
A well-known algorithm to solve this problem is the
Gale-Shapley algorithm, which runs in polynomial time.

It has been proven that stable marriage procedures
can always be manipulated. Whilst the Gale-Shapley
algorithm is computationally easy to manipulate,
we prove that there exist stable marriage
procedures which are NP-hard to manipulate.
We also consider the relationship between voting theory and stable marriage
procedures, showing that voting rules which are NP-hard to manipulate can be
used to define stable marriage procedures which are themselves NP-hard
to manipulate. Finally, we consider the issue that stable marriage
procedures like Gale-Shapley favour one gender over the other,
and we show how to use voting rules to make any stable marriage
procedure gender neutral.


%
%
\end{abstract}

  


\section{Introduction}

The stable marriage problem (SMP) \cite{sm-book} is a well-known problem 
of matching the elements of two sets. 
Given $n$ men and $n$ women, where each person expresses a 
strict ordering over the members of the opposite sex, the problem is  
to match the men to the women 
so that there are no 
two people of opposite sex who would both rather be matched with each other 
than their current partners. 
If there are no such people, all the marriages are said to be {\em stable}. 
Gale and Shapley \cite{gs} proved that 
it is always possible to 
solve the SMP and make all marriages stable, and provided a quadratic
time algorithm which can be used to find
one of two particular but extreme stable marriages,
the so-called {\em male optimal} or {\em female optimal} solution.
%
The Gale-Shapley algorithm 
has been used in many real-life applications, such as in 
systems for matching hospitals to resident doctors \cite{roth-H} and  
the assignment of primary school students in Singapore to secondary 
schools \cite{revisited}. 
Variants of the stable marriage 
problem turn up in many domains. For 
example, the US Navy has a web-based multi-agent system
for assigning sailors to ships \cite{liebowitz}. 

One important issue is whether agents have an incentive
to tell the truth or can manipulate the result by
mis-reporting their preferences. 
Unfortunately, Roth \cite{roth-manip} has proved 
that {\em all} stable marriage procedures
can be manipulated. He demonstrated a stable marriage problem 
with 3 men and 3 women which can be manipulated
{whatever} stable marriage procedure we use. 
This result is in some sense analogous to the classical 
Gibbard Satterthwaite \cite{gibbard,gibbard2}
theorem for voting theory, which states that all voting procedures
are manipulable under modest assumptions provided we have
3 or more voters. 
For voting theory, Bartholdi, Tovey and 
Trick \cite{bartholdi} proposed that computational complexity
might be an escape: whilst manipulation is always
possible, there are voting rules where it is NP-hard
to find a manipulation. 

We might hope that 
computational complexity might also be a barrier
to manipulate stable marriage procedures. 
Unfortunately, the Gale-Shapley algorithm 
is computationally easy to manipulate \cite{revisited}.
We identify here stable marriage procedures that
are NP-hard to manipulate. This can be considered
a first step to understanding if computational
complexity might be a barrier to manipulations. Many questions remain
to be answered. For example, the preferences
met in practice may be highly correlated. Men
may have similar preferences for many of the women. 
Are such profiles computationally difficult to manipulate? 
As a second example, it has been recently recognised 
(see, for example, \cite{average1,average2})
that worst-case
results may represent an insufficient barrier against manipulation 
since they may only apply to problems that are
rare. 
Are there stable marriage procedures which
are difficult to manipulate on average? 

Another drawback of many 
stable marriage procedures such as the one proposed by Gale-Shapley is 
their bias towards one of the two genders. 
The stable matching returned by the Gale-Shapley algorithm
is either male optimal (and
the best possible for every man) but female pessimal
(that is, the worst possible for every woman),
or female optimal but male pessimal. 
It is often desirable to 
use stable marriage procedures that are gender neutral \cite{masarani}.
Such procedures return a stable matching 
that is not affected by swapping the men with the women.
The goal of this paper is to study both the complexity
of manipulation and gender neutrality 
in stable marriage procedures, and to design gender neutral procedures that 
are difficult to manipulate.

It is known that the Gale-Shapley algorithm is computationally
easy to manipulate \cite{revisited}.
Our first contribution is to prove that if the male and female preferences have
a certain form, it is computationally easy to manipulate
{\em any} stable marriage procedure. 
We provide a universal polynomial time manipulation scheme that,
under certain conditions on the preferences, 
guarantees that the manipulator marries his optimal stable
partner irrespective of the stable marriage procedure used. 
On the other hand, our second contribution is to prove that,
when the preferences of the men and women are unrestricted,
there exist stable marriage procedures which are NP-hard to manipulate.

Our third contribution 
is to show that any stable marriage procedure can be made gender neutral
by means of a 
simple pre-processing step which may swap the men with the women. 
This swap can, for instance, be 
decided by a voting rule. 
However, this may give a gender neutral
stable matching procedure which is easy to manipulate.

Our final contribution is a 
stable matching procedure which is both gender neutral and NP-hard to manipulate. 
This procedure uses a voting rule that, considering the male and female
preferences, helps to choose between stable matchings. 
In fact, it picks the stable matching that is most preferred
by the most popular men and women. 
We prove that, if the voting rule used is Single Transferable
Vote (STV) \cite{handbook-sc}, which is NP-hard to manipulate, 
then the resulting stable matching procedure is both gender neutral
and NP-hard to manipulate.
We conjecture that other voting rules which are NP-hard
to manipulate will give rise to stable matching procedures
which are also gender neutral and NP-hard to manipulate. 
Thus, our approach shows how combining voting rules and stable matching 
procedures can be beneficial in two ways: by using preferences 
to discriminate among stable matchings and 
by providing a possible computational shield against manipulation.  

\section{Background}

The {\em stable marriage problem} (SMP) is the problem of finding a 
a matching between the elements of two sets. 
More precisely, given $n$ men and $n$ women, 
where each person strictly orders all members of 
the opposite gender, we wish to marry the men to the women such that there 
are no two people of opposite sex who would both rather 
be married to each other than their current partners. If 
there are no such people, all the marriages are {\em stable}. 

\subsection{The Gale-Shapley algorithm}

The {\em Gale-Shapley algorithm} \cite{gs}
is a well-known algorithm to solve the SMP problem. 
It involves a number of rounds 
where each un-engaged man ``proposes" to his most-preferred woman 
to whom he has not yet proposed. Each woman then considers 
all her suitors and tells the one she most prefers ``maybe" 
and all the rest of them ``No". She is then provisionally ``engaged". 
In each subsequent round, each un-engaged man proposes to one woman 
to whom he has not yet proposed (the woman may or may not already 
be engaged), and the women once again reply with one ``maybe" 
and reject the rest. This may mean that already-engaged 
women can ``trade up", and already-engaged men can be ``jilted".

This algorithm needs a number of steps that is quadratic in $n$, and it guarantees that:
\begin{itemize}
\item  If the number of men and women coincide, and all participants 
express a linear order over all the members of the other group,
everyone gets married. 
Once a woman becomes engaged, 
she is always engaged to someone. So, at the end, there cannot be a 
man and a woman both un-engaged, as he must have proposed to her 
at some point (since a man will eventually propose to every woman, 
if necessary) and, being un-engaged, she would have to have said yes.

\item The marriages are stable. Let Alice be a woman and Bob be a man. 
Suppose they are each married, but not to each other. Upon completion of 
the algorithm, it is not possible for both Alice and Bob to prefer 
each other over their current partners. If Bob prefers Alice 
to his current partner, he must have proposed to Alice before 
he proposed to his current partner. If Alice accepted his proposal, 
yet is not married to him at the end, she must have dumped him 
for someone she likes more, and therefore doesn't like Bob 
more than her current partner. If Alice rejected his proposal, 
she was already with someone she liked more than Bob.
\end{itemize}
%
%
Note that the pairing generated by the Gale-Shapley algorithm is male optimal, 
i.e., every man is paired with his highest ranked feasible partner, and female-pessimal, 
i.e.,  each female is paired with her lowest ranked feasible partner.
It would be the reverse, of course, if the roles of male and female 
participants in the algorithm were interchanged. 

Given $n$ men and $n$ women, a {\em profile} is a sequence
of $2n$ strict total orders, $n$ over the men and $n$ over the women.
In a profile, every woman ranks all the men, and every man ranks all the women.

\begin{example}
\label{ex1}
Assume $n=3$. Let $W=\{w_1,w_2,w_3\}$ and  $M=\{m_1,m_2,m_3\}$
be respectively the set of women and men.
The following sequence of strict total orders defines a profile:
\begin{itemize}
\item $m_1:w_1>w_2>w_3$ (i.e., the man $m_1$ prefers the woman $w_1$ to $w_2$ to $w_3$),
\item $m_2:w_2>w_1>w_3$,
\item $m_3:w_3>w_2>w_1$,
\item $w_1:m_1>m_2>m_3$,
\item $w_2:m_3>m_1>m_2$,
\item $w_3:m_2>m_1>m_3$
\end{itemize}
For this profile, the Gale-Shapley algorithm returns the male optimal solution
$\{(m_1,w_1),(m_2,w_2),(m_3,w_3)\}$. On the other hand, the female optimal solution
is $\{(w_1,m_1),(w_2,m_3),(w_3,$ $m_2)\}$.
\end{example}

\subsection{Gender neutrality and non-manipulability}

A desirable property of a stable marriage procedure is gender neutrality. 
A stable marriage procedure is {\em gender neutral} \cite{masarani} if and only if  when 
we swap the men with the women, we get the same result. 
A related property, called {\em peer indifference} \cite{masarani}, 
holds if the result is not affected by the order in which the members 
of the same sex are considered.
The Gale-Shapley procedure is peer indifferent but it is not gender neutral.
In fact, if we swap men and women in Example \ref{ex1},
we obtain the female optimal solution rather than the male optimal one.
   
Another useful property of a stable marriage procedure is 
its resistance to manipulation. In fact, it would be desirable that 
lying would not lead to better results for the lier.
A stable marriage procedure is {\em manipulable} if
there is a way for one person to mis-report their preferences
and obtain a result which is better than the one 
they would have obtained with their true preferences.

Roth \cite{roth-manip} has proven that stable marriage procedures
can always be {\em manipulated}, 
i.e, that no stable marriage procedures exist which 
always yields a stable outcome and give agents the 
incentive to reveal their true preferences.
He demonstrated a 3 men, 3 women profile which can be manipulated
whatever stable marriage procedure we use. 
%
%
A similar result in a different context is the one by 
Gibbard and Satterthwaite \cite{gibbard,gibbard2}, that
proves that all voting procedures \cite{handbook-sc} are manipulable 
under some modest assumptions. 
In this context, Bartholdi, Tovey and Trick \cite{bartholdi} 
proposed that computational complexity
might be an escape: whilst manipulation is always
possible, there are rules like Single Transferable Vote (STV) 
where it is NP-hard to find a manipulation \cite{stvhard}. 
This resistance to manipulation arises from 
the difficulty of inverting
the voting rule and does not depend on other assumptions like the
difficulty of discovering the preferences of the other voters.
In this paper, we study whether 
computational complexity may also be an escape
from the manipulability of stable marriage procedures. 
Our results are only initial steps to a more complete understanding
of the computational complexity of manipulating 
stable matching procedures. As mentioned before, 
NP-hardness results only address the worst case and may
not apply to preferences met in practice. 

\section{Manipulating stable marriage procedures}
\label{comp1}





A {\em manipulation attempt} by a participant $p$ 
is the mis-reporting of $p$'s preferences.
A manipulation attempt is {\em unsuccessful} 
if the resulting marriage for $p$ is strictly worse 
than the marriage obtained telling the truth.
Otherwise, it is said to be {\em successful}.
A stable marriage procedure is {\em manipulable} if 
there is a profile with a successful manipulation attempt from a participant.



The Gale-Shapley procedure, which depending on how it
is defined returns
either the male optimal or the female optimal solutions,
is computationally easy to manipulate \cite{revisited}.
However, besides these two extreme solutions, 
there may be many other stable matchings. 
Several procedures have been defined 
to return some of these other stable matchings \cite{gusfield}.
Our first contribution is
to show that, under certain conditions on the shape of the male
and female preferences, 
{\em any} stable marriage procedure is computationally 
easy to manipulate. 

Consider a profile $p$ and a woman $w$ in such a profile.
Let $m$ be the male optimal partner for $w$ in $p$, and $n$
be the female optimal partner for $w$ in $p$.
Profile $p$ is said to be  
{\em universally manipulable by $w$}
if the following conditions hold:
\begin{itemize}
\item in the men-proposing Gale-Shapley algorithm, $w$ receives more than one proposal;
\item there exists a woman $v$ such that 
$n$ is the male optimal partner for $v$ in $p$;
\item $v$ prefers $m$ to $n$;
\item $n$'s preferences are $\ldots > v > w > \ldots$; 
\item $m$'s preferences $\ldots w > v > \ldots$.
\end{itemize}

\begin{theorem}
\label{uni}
Consider any stable marriage procedure and any woman $w$.
There is a polynomial manipulation scheme that, 
for any profile which is universally manipulable by $w$, 
produces the female optimal partner for $w$. 
Otherwise, it produces the same partner.
\end{theorem}

\proof{
Consider the manipulation attempt that moves the male optimal partner $m$ of $w$ 
to the lower end of $w$'s preference ordering, obtaining the new profile $p'$.
Consider now the behaviour of the men-proposing Gale-Shapley algorithm on $p$ and $p'$.
Two cases are possible for $p$: 
$w$ is proposed to only by man $m$, 
or it is proposed to also by some other man $o$. In this second case, it must 
be $w$ prefers $m$ to $o$ since
$m$ is the male optimal partner for $w$.

If $w$ is proposed to by $m$ and also by some $o$,
then, when $w$ compares the two proposals, in $p$ she will decide for $m$,
while in $p'$ she will decide for $o$.
At this point, in $p'$, $m$ will have to propose to the next best woman for him, that 
is, $v$, and she will accept because of the assumptions on her preference ordering. 
This means that $n$ (who was married to $v$ in $p$) now in $p'$ has to propose
to his next best choice, that is, $w$, who will accept, since $w$
prefers $n$ to $m$.
So, in $p'$, the male optimal partner for $w$, as well as her 
female optimal partner, is $n$.
This means that there is only one stable partner for $w$ in $p'$.
Therefore, any stable marriage procedure must return $n$ as the partner for $w$.

Thus, if woman $w$ wants to manipulate 
a stable marriage procedure, she can check if the profile is
universally manipulable 
by her. This involves simulating the Gale-Shapley algorithm to see whether 
she is proposed by $m$ only or also by some other man. 
In the former case, she will not do the manipulation.
Otherwise, she will move $m$ to the far right 
it and she will get her female optimal partner, whatever 
stable marriage procedure is used. 
This procedure is polynomial since the Gale-Shapley algorithm takes quadratic time to run. $\Box$
}

\begin{example}
In a setting with 3 men and 3 women, consider 
the profile
$\{ m_1:w_1>w_2>w_3; 
    \mbox{ } m_2:w_2>w_1>w_3;
    \mbox{ } m_3:w_1>w_2>w_3;\}$ $\{ 
    w_1:m_2>m_1>m_3;
    \mbox{ } w_2:m_1>m_2>m_3;
    \mbox{ } w_3:m_1>m_2>m_3;\}$
In this profile, the male optimal solution is $\{(m_1,w_1),(m_2,$ $w_2),(m_3,w_3)\}.$
This profile is universally manipulable by $w_1$. 
In fact, woman $w_1$ can successfully manipulate by moving $m_1$ after $m_3$, and obtaining
the marriage $(m_2,w_1)$, thus getting her female optimal partner.
Notice that this holds no matter what stable marriage procedure is used.
This same profile is not universally manipulable by $w_2$ or $w_3$,
since they receive just one proposal in the men-proposing Gale-Shapley algorithm.
In fact, woman $w_2$ cannot manipulate: trying to move $m_2$ after $m_3$ gets a
worse result. Also, woman $w_3$ cannot manipulate since 
her male optimal partner is her least preferred man. 
\end{example}

Restricting to universally manipulable profiles 
makes manipulation computationally easy. On the
other hand, if we allow all possible profiles, 
there are stable marriage procedures that are 
NP-hard to manipulate. 
The intuition is simple.  
We construct a stable marriage procedure that is 
computationally easy to compute but NP-hard to invert. 
To manipulate, a man or a woman will essentially need to be able to 
invert the procedure to choose between the exponential 
number of possible preference orderings. Hence,
the constructed stable marriage procedure will be NP-hard
to manipulate. The stable marriage procedure used
in this proof is somewhat ``artificial''. However, 
we will later propose a stable marriage
procedure which is more natural while remaining NP-hard
to manipulate. This procedure 
selects the stable matching that is most preferred
by the most popular men and women. It
is an interesting open question to devise
other stable marriage procedures which are
``natural'' and computationally difficult to manipulate.

\begin{theorem}
\label{np}
There exist stable marriage procedures
for which deciding the existence of a successful manipulation is NP-complete. 
\end{theorem}

\proof{
We construct a stable marriage procedure which chooses between
the male and female optimal solution
based on whether the profile encodes
a NP-complete problem and its polynomial witness.
The manipulator's preferences define 
the witness. The other people's preferences 
define the NP-complete problem. Hence, the 
manipulator needs to be able to solve
a NP-complete problem to be able to
manipulate successfully. 
Deciding if there is a successful manipulation for 
this stable marriage procedure is clearly in NP 
since we can compute male and female optimal
solutions in polynomial time, and we can
check a witness to a NP-complete problem
also in polynomial time. 

Our stable marriage procedure
is defined to work on $n+3$ men ($m_1$, $m_2$
and $p_1$ to $p_{n+1}$) and $n+3$ women 
($w_1$, $w_2$ and $v_1$ to $v_{n+1}$). 
It returns the female optimal
solution if the preferences of woman $w_1$ encode 
a Hamiltonian path in a directed graph encoded by the
other women's preferences, otherwise
it returns the male optimal solution. 
The $3$rd to $n+2$th preferences of woman $w_1$
encode a possible Hamiltonian path in a $n$ node graph. In
particular, if the
$2+i$th man in the preference ordering 
of woman $w_1$ for $i>0$ is man
$p_{j}$, then the path goes from vertex $i$
to vertex $j$. The preferences of the women 
$v_{i}$ for $i \leq n$ encode the graph in which we find
this Hamiltonian path. In particular, if 
man $p_{j}$ for $j<n+1$ and $j\neq i$
appears before 
man $p_{n+1}$ in the preference list of woman $w_{i}$, then
there is a directed edge in the graph from $i$ to $j$. 
It should be noticed that any graph 
can be produced using this construction.

Given a graph which is not complete
in which we wish to find
a Hamiltonian path, we now build a special 
profile. Woman $w_1$ will be able to manipulate this profile
successfully iff the graph
contains a Hamiltonian path.
In the profile, woman $w_1$ most prefers to marry man $m_{1}$
and then man $m_{2}$. Consider any pair of vertices $(i,j)$
not in the graph. Woman $w_1$ puts man $p_j$ at position
$2+i$ in her preference order. She puts all other
$p_j$'s in any arbitrary order. This construction will guarantee
that the preferences of $w_1$ do not represent
a Hamiltonian path. Woman $w_2$ most prefers to 
marry man $m_{2}$. 
Woman $v_{i}$ most
prefers to marry man $p_{i}$, 
and has preferences for the other men $p_j$ 
according to the edges from vertex $i$. 
Man $m_1$ most prefers woman $w_2$.
Man $m_2$ most prefers woman $w_1$. 
Finally, man $p_i$ most prefers woman $v_i$. 
All other unspecified preferences can be chosen in
any way. By construction, all first choices
are different.
Hence, the male optimal solution has the men married to their
first choice, whilst the 
female optimal solution has the women married to their
first choice.

The male optimal solution has woman $w_1$ married to man $m_2$.
The female optimal solution has woman $w_1$ married to man $m_1$.
By construction, the preferences of woman $w_1$ do not
represent a Hamiltonian path. Hence our stable matching 
procedure returns the male optimal solution: woman
$w_1$ married to man $m_2$. The only successful
manipulation then for woman $w_1$ is if she can
marry her most preferred choice, man $m_1$.
As all first choices are different, 
woman $w_1$ cannot successfully manipulate
the male or female optimal solution. 
Therefore, she must manipulate her preferences
so that she spells out a Hamiltonian path 
in her preference ordering, and our stable marriage procedure 
therefore returns the female optimal solution.
This means she can successful
manipulate iff there is a Hamiltonian path. Hence, 
deciding if there is a successful manipulation is
NP-complete. $\Box$} 

Note that we can modify the proof by
introducing $O(n^2)$ men so that the
graph is encoded in the tail of the preferences
of woman $w_2$. This means that it remains
NP-hard to manipulate a stable marriage
procedure even if we collude with all but one of the women. 
It also means that it is NP-hard to manipulate
a stable marriage procedure when the problem 
is imbalanced and there are just 2 women
but an arbitrary number of men. 
Notice that this procedure is not peer indifferent, since it gives
special roles to different men and women.
However, it is possible to make it peer indifferent, 
so that it computes the same result if we rename the men
and women. For instance, we just take the men's
preferences and compute from them a total ordering
of the women (e.g. by running an election with these
preferences). Similarly, we take the women's 
preferences and compute from them a total ordering
of the men. We can then use these orderings to assign
indices to men and women. 
Notice also this procedure is not gender neutral. If 
we swap men and women, we may get a different result.
We can, however, use the simple procedure proposed
in the next section to make it gender neutral. 

\section{Gender neutrality}

As mentioned before, a weakness of many stable marriage procedures
like the Gale-Shapley procedure and the procedure presented in 
the previous section, is that they are not gender neutral.
They may greatly favour one sex over the other. 
We now present a simple and universal technique 
for taking any stable marriage procedure
and making it gender neutral.
We will assume that the men and the women are named
from 1 to $n$. 
We will also say that the men's preferences are {\em isomorphic}
to the women's preferences iff there is a bijection between
the men and women that preserves both the men's and women's
preferences. In this case, it is easy 
to see that there is only one stable matching.


We can convert any 
stable marriage procedure into one that is gender neutral by
adding a pre-round in which we choose if we swap the men with the women.
The idea of using pre-rounds for enforcing certain 
properties is not new and has been used for example in \cite{preround} to 
make manipulation of voting rules NP-hard. 
The goal of our pre-round is, instead, to ensure gender-neutrality.
More precisely, 
for each gender we compute its {\em signature}: a vector of numbers constructed
by concatenating together each of the individual preference lists. 
Among all such vectors, the signature is 
the lexicographically smallest vector under reordering
of the members of the chosen gender and renumbering of the members of the other gender.

\begin{example}
Consider the following profile with 3 men and 3 women.
$\{           m_1:w_2>w_1>w_3;
    \mbox{ }  m_2:w_3>w_2>w_1;
    \mbox{ }  m_3:w_2>w_1>w_3 \}$
             $\{ w_1:m_1>m_2>m_3;
     \mbox{ } w_2:m_3>m_1>m_2;
     \mbox{ } w_3:m_2>m_1>m_3\}.$
The signature of the men is 123123312: each group of three digits represents 
the preference ordering of a man; men $m_2$ and $m_3$ and
women $w_1$ and $w_2$ have been swapped with each other to obtain
the lexicographically smallest vector.
The signature of the women is instead 123213312. 
\end{example}

Note that this vector can be computed in $O(n^2)$ time. For each
man, we put his preference list first, then reorder
the women so that this man's preference list reads $1$ to $n$. Finally,
we concatenate the other men's preference lists in lexicographical
order. We define the signature as the smallest such vector.

Before applying any stable marriage procedure, we propose to 
pre-process the profile according to the following rule, 
that we will call {\em gn-rule}
(for gender neutral):
If the male signature is smaller than the female
signature, then we swap the men with the women
before calling the stable marriage procedure.
On the other hand, if the male signature is equal or greater than
the female signature, we will not
swap the men with the women before calling the stable marriage procedure.
In the example above, the male signature is smaller than the female signature, thus 
men and women must be swapped before using the stable marriage procedure.

\begin{theorem}
Consider any 
stable marriage procedure, say $\mu$.
Given a profile $p$, consider the new procedure $\mu'$ obtained by 
applying $\mu$ to gn-rule(p). This new procedure returns a stable marriage and 
it is gender neutral.
Moreover, if $\mu$ is peer indifferent, then $\mu'$ is peer indifferent as well.
\end{theorem}

\proof{
To prove gender neutrality, we consider three cases:
\begin{itemize}
\item If the male signature is smaller than the female signature,
the gn-rule swaps the men with the women.
Thus we would apply $\mu$ to swapped genders.

To prove that the new procedure is gender neutral, we must prove that, if 
we swap the men with the women, the result is the same.
If we do this swap, their signatures will be swapped.
Thus the male signature will result larger than the female signature, and therefore the gn-rule
will not swap men and women.
Thus procedure $\mu$ will be applied to swapped genders.

\item If the male signature is larger than the female signature,
the gn-rule leaves the profile as it is.
Thus $\mu$ is applied to profile $p$.

If we swap the genders, 
the male signature will result smaller than the female signature, 
and therefore the gn-rule will perform the swap.
Thus procedure $\mu$ will be applied to the original profile $p$.

\item 
If the male and female signatures are identical, 
the men and women's preferences are isomorphic and there is only one stable matching.
Any stable marriage procedure must therefore return this matching, 
and hence it is gender neutral.
\end{itemize}

As for peer indifference, 
if we start from a profile obtained by reordering men or women,
the signatures will be the same and thus the gn-rule will perform the same 
(either swapping or not). Thus the result of applying the whole procedure to the 
reordered profile will be the same as the one obtained by using the given profile. $\Box$
}

If we are not concerned about preserving peer indifference,
or if we start from a non-peer indifferent matching procedure, 
we can use a much simpler version of the gn-rule, where 
the signatures are obtained directly from the profile without 
considering any reordering/renaming of men or women.
This simpler approach is still sufficient to guarantee gender neutrality, but 
might produce a procedure which is not peer indifferent.


\section{Voting Rules and Stable \\ Marriage Procedures}

We will now see how we can exploit results about voting rules
to build stable marriage procedures which are both gender neutral
and difficult to manipulate.

\subsection{A score-based matching procedure: 
gender neutral but easy to manipulate}

Given a profile, consider a set of its stable matchings. For simplicity,
consider the set containing only the male and female optimal stable matchings.
However, there is no reason why we could not consider a larger 
polynomial size
set. 
For example, we might consider all stable matchings found
on a path through the stable marriage lattice \cite{lattice}
between the male and female optimal, 
or we may simply run twice any procedure 
computing a set of stable marriages, swapping genders the second time.
We can now use the men and women's preferences to rank 
stable matchings in the considered set.
For example, as in \cite{ilgjacm87},
we can score a matching as the sum of the
men's ranks of their partners and of the women's ranks
of their partners. 

We then choose between the stable matchings in our given set according
to which has the smallest score.
Since our set contains only the male and the female optimal matches,
we choose between the male and female optimal stable matchings according 
to which has the lowest score. 
If the male optimal and the female optimal stable matching  
have the same score, we use the signature of men and women, as 
defined in the previous section, to tie-break.
It is possible to show that the resulting matching procedure, 
which returns the male optimal or the female optimal stable matching according 
to the scoring rule (or, if they have the same score, according to the signature)
is gender neutral. 

Unfortunately, this procedure
is easy to manipulate. For a man, it is sufficient to 
place his male optimal partner in
first place in his preference list, and his female optimal partner in last
place. If this manipulation does not give the man his male
optimal partner, then there is no manipulation that will.
A woman manipulates the result in a symmetric way.

\subsection{Lexicographical minimal regret}

Let us now consider a more complex score-based  matching procedure
to choose between two (or more) stable matchings which will be computationally
difficult to manipulate. The intuition behind the procedure
is to choose between stable matchings according to the
preferences of the most preferred men or women. 
In particular, we will pick the stable matching that is most preferred
by the most popular men and women. 
Given a voting rule, we order the men
using the women's preferences and order the women using the men's preferences.
We then construct a male score vector for a
matching using this ordering of the men (where a more preferred man 
is before a less preferred one).
The $i$th element of the
male score vector is the integer $j$ iff the $i$th man in
this order is married to his $j$th most preferred
woman. A large male score vector
is a measure of dissatisfaction with the matching
from the perspective of the more preferred men.
A female score vector is computed in an analogous manner. 

The overall score
for a matching is the lexicographically largest of its male and female
score vectors. A large overall score
corresponds to dissatisfaction with the matching
from the perspective of the more preferred men or women.
We then choose the stable matching from
our given set which has the lexicographically least overall score.
That is, we choose the stable matching which carries
less regret for the more preferred men and women.

In the event of a tie, we can use any
gender neutral tie-breaking procedure, 
such as the one based on signatures described above. 
Let us call this procedure the {\em lexicographical minimal regret} 
stable marriage procedure.
In particular, when voting rule $v$ is used to order the men and women we 
will call it a \emph{v-based} lexicographical minimal regret 
stable marriage procedure.
It is easy to see that this procedure is gender neutral.
In addition, it is computationally hard
to manipulate. Here we consider using STV \cite{handbook-sc}
to order the men and women. However, we conjecture
that similar results will hold for stable matching procedures
which are derived from other voting rules which are NP-hard to manipulate.

In the STV rule each voter provides
a total order on candidates and, initially, an individual's
vote is allocated to his most preferred candidate. The
quota of the election is the minimum number of votes necessary
to get elected. If no candidate exceeds the quota, then, the candidate
with the fewest votes is eliminated, and his votes are equally
distributed among the second choices of the voters who had
selected him as first choice. This step is repeated until some 
candidate exceeds the quota. In the following theorem we assume a 
quota of at least half of the number of voters.

\begin{theorem}
It is NP-complete to decide if an agent can manipulate the
STV-based lexicographical minimal regret stable marriage procedure.
\end{theorem}

\proof{
We adapt the reduction used to prove that
constructive manipulation of the STV rule
by a single voter is NP-hard \cite{stvhard}.
In our proof, we need to consider how the STV rule
treats ties. 
For example, ties will occur among all men and all women, since 
we will build a profile 
where every man and every woman have different first choice.
Thus STV will need to tie break between
all the men (and between all the women).
We suppose that in any such tie break,
the candidate alphabetically last
is eliminated. We also suppose that
a man $h$ will try to manipulate
the stable marriage procedure by mis-reporting
his preferences.

To prove membership in NP, we observe
that a manipulation is a polynomial witness.
To prove NP-hardness, we give a reduction
from 3-COVER. Given a  set $S$ with $|S|=n$,
subsets and subsets $S_i$ with $i \in [1,m]$, $|S_i=3|$
and $S_i \subset S$, we ask if there
exists an index set $I$ with $|I|=n/3$ and
$\bigcup_{i \in I} S_i =S$.

We will construct a profile of preferences for the men so that
the only possibility is for STV to order
first one of only two women, $w$ or $y$.
The manipulator $h$ will try to vote strategically so that
woman $y$ is ordered first. This will
have the consequence that
we return the male optimal stable marriage in which the
manipulator marries his first choice $z_1$.
On the other hand, if $w$ is ordered
first, we will return the female optimal stable
marriage in which the manipulator is married to his second
choice $z_2$. 

The following sets of women participate
in the problem:
\begin{itemize}
\item two possible winners of the first STV election,
$w$ and $y$;
\item   $z_1$ and $z_2$ who are the first
two choices of the manipulator;
\item ``first losers'' in this election, $a_i$ and $b_i$ for $i \in [1,m]$;
\item ``second line'' in this election, $c_i$ and $d_i$ for $i \in [1,m]$;
\item ``$e$-bloc'', $e_i$ for $i \in [0,n]$;
\item ``garbage collectors'', $g_i$ for $i \in [1,m]$;
\item ``dummy women'', $z_{i,j,k}$ where $i \in [1,19]$ and
$j$ and $k$ depend on $i$ as outlined in
the description given shortly for the men's preferences
(e.g. for $i=1$, $j=1$ and $k \in [1,12m-1]$ but
for $i \in [6,8]$, $j \in [1,m]$ and $k \in [1,6m+4j-6]$). 
\end{itemize}
Ignoring the manipulator, the men's preferences
will be constructed so that $z_1$, $z_2$ and
the dummy women are the first women
eliminated by the STV rule,
and that $a_i$ and $b_i$ are
$2m$ out of the next $3m$ woman eliminated.
In addition, let $I = \{ i : b_i \ is \ eliminated \
before \ a_i\}$. Then the men's preferences
will be constructed so that STV orders woman $y$ first
if and only if $I$ is a 3-COVER.
The manipulator
can ensure $b_i$ is eliminated by the STV
rule before $a_i$
for $i \in I$ by placing $a_i$ in the $i+1$th position
and $b_i$ otherwise.

The men's preferences are constructed as follows
(where preferences are left unspecified, they can
be completed in any order):
\begin{itemize}
\item a man $n$ with preference $(y, \ldots)$ and
$\forall k \in [1,12m-1]$ a man
with 
$(z_{1,1,k},y,\ldots)$;
\item a man $p$ with preference $(w,y,\ldots)$ and
$\forall k \in [1,12m-2]$ a man
with 
$(z_{2,1,k},w,y,\ldots)$;
\item a man $q$ with preference $(e_0,w,y,\ldots)$ and
$\forall k \in [1,10m+2n/3-1]$ a man
with 
$(z_{3,1,k},e_0,w,y,\ldots)$;
\item $\forall j \in [1,n]$,
a man with preference $(e_j,w,y,\ldots)$ and
$\forall k \in [1,12m-3]$
a man with preference 
$(z_{4,j,k},e_j,w,y,$ $\ldots)$;
\item $\forall j \in [1,m]$,
a man $r_j$ with preference $(g_j,w,y,\ldots)$ and
$\forall k \in [1,12m-1]$
a man with preference 
$(z_{5,j,k},g_j,$ $w,y,\ldots)$;
\item $\forall j \in [1,m]$,
a man with preference $(c_j,d_j,w,y,\ldots)$
and $\forall k \in [1,6m+4j-6]$ a
man with preference 
$(z_{6,j,k},c_j,d_j,$ $w,y,\ldots)$,
and for each of the three $k$ s.t. $k \in S_j$,
a man with preference 
$(z_{7,j,k},c_j,e_k,w,y,\ldots)$,
and one with preference 
$(z_{8,j,k},c_j,e_k,w,y,\ldots)$;
\item $\forall j \in [1,m]$,
a man with preference $(d_j,c_j,w,y,\ldots)$
and $\forall k \in [1,6m+4j-2]$ a
man with preference 
$(z_{9,j,k},d_j,c_j,$ $w,y,\ldots)$,
one with preference 
$(z_{10,j,k},d_j,e_0,$ $w,y,\ldots)$,
and one with 
$(z_{11,j,k},d_j,e_0,w,y,\ldots)$;
\item $\forall j \in [1,m]$,
a man with preference $(a_j,g_j,w,y,\ldots)$
and $\forall k \in [1,6m+4j-4]$
a man with preference 
$(z_{12,j,k},a_j,g_j,$ $w,y,\ldots)$,
one with preference 
$(z_{13,j,k},a_j,$ $c_j,$ $w,y,\ldots)$,
one with preference 
$(z_{14,j,k},a_j,b_j,w,y,\ldots)$,
and one with preference 
$(z_{15,j,k},a_j,b_j,w,y,\ldots)$.
\item $\forall j \in [1,m]$,
a man with preference $(b_j,g_j,w,y,\ldots)$
and $\forall k \in [1,6m+4j-4]$
a man with preference 
$(z_{16,j,k},b_j,g_j,$ $w,y,\ldots)$,
one with preference 
$(z_{17,j,k},b_j,$ $d_j,$ $w,y,\ldots)$,
one with preference 
$(z_{18,j,k},b_j,a_j,w,y,\ldots)$,
and one with preference 
$(z_{19,j,k},b_j,a_j,w,y,\ldots)$.
\end{itemize}

Note that each woman is ranked first by exactly one man.
The women's preference will be set
up so that the manipulator $h$ is assured at least that he
will marry his second choice, $z_2$ 
as this will be his female optimal partner. 
To manipulate the election, the manipulator needs to put $z_1$ first 
in his preferences and to report the rest
of his preferences so that the result returned is the male optimal 
solution. As all woman are ranked first by exactly one
man, the male optimal matching marries $h$ with $z_1$. 

When we use STV to order the women,
$z_1$, $z_2$ and $z_{i,j,k}$ are
alphabetically last so are eliminated
first by the tie-breaking rule. This leaves the following profile:
\begin{itemize}
\item $12m$ men with preference $(y, \ldots)$;
\item $12m-1$ men with preference $(w,y,\ldots)$;
\item $10m+2n/3$ men with preference $(e_0,w,y,\ldots)$;
\item $\forall j \in [1,n]$,
$12m-2$ men with preference $(e_j,w,y,$ $\ldots)$;
\item $\forall j \in [1,m]$,
$12m$ men with preference $(g_j,w,y,$ $\ldots)$;
\item $\forall j \in [1,m]$,
$6m+4j-5$ men with preference $(c_j,d_j,w,y,$ $\ldots)$,
and for each of the three $k$ such that $k \in S_j$,
two men with preference $(c_j,e_k,w,y,\ldots)$;
\item $\forall j \in [1,m]$,
$6m+4j-1$ men with preference $(d_j,c_j,w,y,$ $\ldots)$,
and two men with preference $(d_j,e_0,w,y,\ldots)$,
\item $\forall j \in [1,m]$,
$6m+4j-3$ men with preference $(a_j,g_j,w,y,$ $\ldots)$,
a man with preference $(a_j,c_j,w,y,\ldots)$,
and two men with preference $(a_j,b_j,w,y,\ldots)$;
\item $\forall j \in [1,m]$,
$6m+4j-3$ men with preference $(b_j,g_j,w,y,$ $\ldots)$
a man with preference $(b_j,d_j,w,y,\ldots)$,
and two men with preference $(b_j,a_j,w,y,\ldots)$.
\end{itemize}

At this point, the votes are identical (up to renaming
of the men) to the profile constructed in the proof of Theorem 1
in \cite{stvhard}.
Using the same argument as there, it follows that the manipulator
can ensure that STV orders woman
$y$ first instead of $w$
if and only if
there is a 3-COVER.
The manipulation
will place $z_1$ first in $h$'s preferences. 
Similar to the proof of
Theorem 1 in \cite{stvhard},
the manipulation puts woman $a_j$ in $j+1$th place and $b_j$ otherwise
where $j \in J$ and $J$ is any index set of a 3-COVER.

The women's preferences are as follows:
\begin{itemize}
\item the woman $y$ with preference $(n, \ldots)$;
\item the woman $w$ with preference $(q,\ldots)$;
\item the woman $z_1$ with preference $(p,\ldots)$;
\item the woman $z_2$ with preference $(h,\ldots)$;
\item the women $g_i$ with preference $(r_i,\ldots)$;
\item the other women with any preferences which are
first-different, and which ensure STV orders $r_0$ first
and $r_1$ second overall.
\end{itemize}

Each man is ranked first by exactly one woman.
Hence, the female optimal stable matching is
the first choice of the women.
The male score vector of the male optimal stable
matching is $(1,1,\ldots,1)$. Hence, the
overall score vector of the male optimal
stable matching equals the female score vector
of the male optimal stable matching.
This is $(1,2,\ldots)$ if the manipulation is
successful and $(2,1,\ldots)$ if it is not.
Similarly, the overall score vector of the
female optimal stable matching equals the
male score vector of the female optimal stable matching.
This is $(1,3,\ldots)$. Hence the lexicographical
minimal regret stable marriage procedure will return
the male optimal stable matching iff there is
a successful manipulation of the STV rule.
Note that the profile used in this proof 
is not universally
manipulable. The first choices of the man are
all different and each woman therefore only receives one
proposal in the men-proposing Gale-Shapley algorithm. $\Box$ 
}

We can thus see how the proposed matching procedure is 
reasonable and appealing. In fact,  
it allows to discriminate among stable matchings according 
to the men and women's preferences and it is difficult to manipulate
while ensuring gender neutrality.

\section{Related work}


In \cite{masarani} fairness of a matching procedure is defined in terms 
of four axioms, two of which are gender neutrality and peer indifference. 
Then, the existence of a matching procedures satisfying all or a 
subset of the axioms is considered in terms of restrictions on preference 
orderings.
Here, instead, we propose a preprocessing step that allows 
to obtain a gender neutral matching procedure from any matching procedure 
without imposing any restrictions on the preferences in the input. 
   
A detailed description of results about manipulation
of stable marriage procedures can be found in \cite{huang}.  
In particular, several early results \cite{demange,dubins, gale,roth-manip} 
indicated the futility of men lying, focusing later work mostly on strategies in which 
the women lie. 
Gale and Sotomayor \cite{gale2} presented the manipulation strategy in 
which women truncate their
preference lists. Roth and Vate \cite{roth3}
discussed strategic issues when the stable matching is chosen at random, 
proposed
a truncation strategy and showed that every stable matching can be
achieved as an equilibrium in truncation strategies.
We instead do not allow the elimination of men from a woman's 
preference ordering, but permit reordering of the preference lists.


Teo et al. \cite{revisited} suggested lying strategies for an individual
woman, and proposed an algorithm to find the best partner with the male optimal procedure.
We instead focus on the complexity of determining if 
the procedure can be manipulated to obtain a better result. 
Moreover, we also provide a universal manipulation scheme that,
under certain conditions on the profile, assures that the female optimal 
partner is returned.


Coalition manipulation is considered in 
\cite{huang}. Huang 
shows how a coalition of men can get a better result in the men-proposing Gale-Shapley 
algorithm. By contrast, we do not consider a
coalition but just a single manipulator, and do
not consider just the Gale-Shapley algorithm. 

\section{Conclusions}

We have studied the manipulability and gender neutrality of stable
marriage procedures. We first looked at whether, as with voting rules, 
computationally complexity might be a barrier to manipulation. 
It was known already that one prominent stable
marriage procedure, the Gale-Shapley algorithm, is computationally easy to manipulate. 
We proved that, under some simple restrictions on agents' preferences,
{\em all} stable marriage procedures are in fact easy to manipulate. 
Our proof provides an universal manipulation which an agent
can use to improve his result. 
On the other hand, when preferences are unrestricted,
we proved that there exist stable marriage procedures which
are NP-hard to manipulate. We also showed how to 
use a voting rule to choose between stable
matchings. In particular, we gave a stable
marriage procedure which picks
the stable matching that is most preferred
by the most popular men and women. 
This procedure inherits the computational
complexity of the underlying voting rule. Thus, when
the STV voting rule (which is NP-hard to 
manipulate) is used to compute the most
popular men and women, the corresponding stable marriage procedure
is NP-hard to manipulate. Another desirable property of stable
marriage procedures is gender neutrality. 
Our procedure of turning a voting rule
into a stable marriage procedure is gender neutral. 

This study of stable marriage procedures is
only an initial step to understanding if computational
complexity might be a barrier to manipulation. 
Many questions remain
to be answered. For example, if preferences are correlated,
are stable marriage procedures still computationally 
hard to manipulate? As a second example, are there stable marriage 
procedures which are difficult to manipulate on average? 
There are also many interesting and related questions connected
with privacy and mechanism design. For instance,
how do we design a decentralised stable marriage procedure
which is resistant to manipulation and in which 
the agents do not share their preference lists?
As a second example, how can side payments 
be used in stable marriage procedures to prevent 
manipulation?

\bibliographystyle{abbrv}

\begin{thebibliography}{1}


\bibitem{handbook-sc}
K.J.~Arrow, A.K.~Sen and K.~Suzumura.
Handbook of Social Choice and Welfare.
North Holland, Elsevier, 2002


\bibitem{stvhard}
J.~Bartholdi and J.~Orlin. 
Single transferable vote resists strategic voting. 
In \emph{Social Choice and Welfare, 8(4):341-354}, 1991.

\bibitem{bartholdi}
J.J.~Bartholdi, C.~A.~Tovey and M.~A.~Trick,
The computational difficulty of manipulating an election. 
In \emph{Social Choice and Welfare 6(3):227-241}, 1989.


\bibitem{average2}
V. Conitzer and T. Sandholm. 
Nonexistence of voting rules that are usually hard to manipulate. 
In{\em  Proc. AAAI'06}, 2006.

\bibitem{preround}
V. Conitzer and T. Sandholm. 
Universal Voting Protocol Tweaks to Make Manipulation Hard. 
In {\em Proc. IJCAI'03}: 781-788, 2003.


\bibitem{demange}
G.~Demange, D.~Gale, and M.~Sotomayor. A further note on the
stable matching problem. In \emph{Discrete Applied Mathematics, 16:217-222}, 1987.

\bibitem{dubins} 
L.~Dubins and D.~Freedman. Machiavelli and the Gale-Shapley algorithm.
In \emph{American Mathematical Monthly, 88:485-494}, 1981.

\bibitem{gs} 
D.~Gale, L.~S.~Shapley. 
College Admissions and the Stability of Marriage. 
In \emph{Amer. Math. Monthly, 69:9-14}, 1962.


\bibitem{gale}
D.~Gale and M.~Sotomayor. Some remarks on the stable matching prob-
lem. \emph{Discrete Applied Mathematics, 11:223-232}, 1985.

\bibitem{gale2}
D.~Gale and M.~Sotomayor. Machiavelli and the stable matching
problem. \emph{American Mathematical Monthly, 92:261-268}, 1985.



\bibitem{gibbard}
A.~Gibbard. 
Manipulation of Voting Schemes: A General Result.
In \emph{Econometrica, 41(3):587-601}, 1973.


\bibitem{sm-book}
D.~Gusfield and R.~W.~Irving.  
The Stable Marriage Problem: Structure and Algorithms.
MIT Press, Boston, Mass.,1989.

\bibitem{gusfield}
D.~Gusfield.
Three fast algorithms for four problems in stable marriage.
In \emph{SIAM J. of Computing, 16(1)}, 1987.


\bibitem{huang}
C.-C.~Huang. Cheating by men in the Gale-Shapley stable matching algorithm. 
In \emph{ESA'06}, pages 418-431,
Springer-Verlag, 2006.


\bibitem{ilgjacm87}
Robert W.~Irving, P.~Leather and D.~Gusfield,
An efficient algorithm for the ``optimal'' stable marriage,
In \emph{ JACM 34(3):532-543},1987.


\bibitem{lattice}
B.~Klaus and F.~Klijin.
Procedurally fair and stable matching.
In \emph{Economic Theory, 27:431-447}, 2006.



\bibitem{liebowitz}
J. Liebowitz and J. Simien.
Computational efficiencies for multi-agents: a 
look at a multi-agent system for sailor assignment. 
In \emph{Electronic government: an International Journal 2 (4):384-402}, 2005.

\bibitem{masarani}
F.~Masarani and S.S.~Gokturk. 
On the existence of fair matching algorithms. 
In \emph{ Theory and Decision, 26:305-322}, Kluwer, 1989.

\bibitem{average1}
A.D.~Procaccia and J.S.~Rosenschein.
Junta Distributions and the Average-Case Complexity of Manipulating Elections.
In {\it JAIR 28:157-181}, 2007. 


 
\bibitem{roth-manip}
A.~E.~Roth. 
The Economics of Matching: Stability and Incentives. 
In \emph{Mathematics of Operations Research, 7:617-628}, 1982.


\bibitem{roth-H}
A.~E.~Roth.
The evolution of the labor market for medical interns and residents: a case study in game theory.
In \emph {Journal of Political Economy 92:991-1016}, 1984. 

\bibitem{roth2}
A.~Roth and M.~Sotomayor. Two-sided matching: A study in game-theoretic modeling and analysis. \emph{Cambridge University Press}, 1990.

\bibitem{roth3}
A.~Roth and V.~Vate. Incentives in two-sided matching with random stable
mechanisms. In \emph{Economic Theory, 1(1):31-44}, 1991.


\bibitem{gibbard2}
M.~A.~Satterthwaite. 
Strategy-proofness and Arrow's conditions: Existence and correspondence theorems for voting procedures and social welfare function. 
In \emph{Economic Theory, 10(3):187-217}, 1975.


\bibitem{revisited}
C.-P.~Teo, J.~Sethuraman and W.-P.~Tan. 
Gale-Shapley Stable Marriage Problem Revisited: Strategic Issues and Applications. In \emph{Management Science, 47(9):1252-1267}, 2001.


















\end{thebibliography}

\end{document}